\documentclass[sn-mathphys,Numbered]{sn-jnl}

\usepackage{graphicx}%
\usepackage{multirow}%
\usepackage{amsmath,amssymb,amsfonts}%
\usepackage{amsthm}%
\usepackage{mathrsfs}%
\usepackage[title]{appendix}%
\usepackage{xcolor}%
\usepackage{textcomp}%
\usepackage{manyfoot}%
\usepackage{booktabs}%
\usepackage{algorithm}%
\usepackage{algorithmicx}%
\usepackage{algpseudocode}%
\usepackage{listings}%
\usepackage{hyperref}
\usepackage{float}
\usepackage{verbatim} 
\usepackage{comment}
\usepackage{caption}
\usepackage{subcaption}
\usepackage{url}
\usepackage{soul}

\begin{document}

\title[VPIT: Real-time Embedded Single Object 3D Tracking Using Voxel Pseudo Images]{VPIT: Real-time Embedded Single Object 3D Tracking Using Voxel Pseudo Images}


\author*[1]{\fnm{Illia} \sur{Oleksiienko}}\email{io@ece.au.dk}

\author[2]{\fnm{Paraskevi} \sur{Nousi}}\email{paranous@csd.auth.gr}

\author[3,2]{\fnm{Nikolaos} \sur{Passalis}}\email{passalis@auth.gr}

\author[2]{\fnm{Anastasios} \sur{Tefas}}\email{tefas@csd.auth.gr}

\author[1]{\fnm{Alexandros} \sur{Iosifidis}}\email{ai@ece.au.dk}

\affil*[1]{\orgdiv{Department of Electrical and Computer Engineering}, \orgname{Aarhus University}, \orgaddress{\street{Finlandsgade}, \city{Aarhus}, \postcode{8200}, \country{Denmark}}}

\affil[2]{\orgdiv{Department of Informatics}, \orgname{Aristotle University of Thessaloniki}, \orgaddress{\street{University Campus}, \city{Thessaloniki}, \postcode{54124}, \country{Greece}}}

\affil[3]{\orgdiv{Department of Chemical Engineering}, \orgname{Aristotle University of Thessaloniki}, \orgaddress{\street{University Campus}, \city{Thessaloniki}, \postcode{54124}, \country{Greece}}}


\abstract{
In this paper, we propose a novel voxel-based 3D single object tracking (3D SOT) method called Voxel Pseudo Image Tracking (VPIT). VPIT is the first method that uses voxel pseudo images for 3D SOT. The input point cloud is structured by pillar-based voxelization, and the resulting pseudo image is used as an input to a 2D-like Siamese SOT method. 
The pseudo image is created in the Bird's-eye View (BEV) coordinates, and therefore the objects in it have constant size. Thus, only the object rotation can change in the new coordinate system and not the object scale. For this reason, we replace multi-scale search with a multi-rotation search, where differently rotated search regions are compared against a single target representation to predict both position and rotation of the object. 
Experiments on KITTI \cite{2012kitti} Tracking dataset show that VPIT is the fastest 3D SOT method and maintains competitive Success and Precision values.
Application of a SOT method in a real-world scenario meets with limitations such as lower computational capabilities of embedded devices and a latency-unforgiving environment, where the method is forced to skip certain data frames if the inference speed is not high enough.
We implement a real-time evaluation protocol and show that other methods lose most of their performance on embedded devices, while VPIT maintains its ability to track the object.
}





\maketitle

\section{Introduction}
With the rise of robotics usage in real-world scenarios, there is a need to develop methods for understanding the 3D world in order to allow a robot to interact with objects. To this end, efficient methods for 3D object detection, tracking, and active perception are needed. Such methods provide the main source of information for scene understanding, as having high-quality object detection and tracking outputs increases the chances for a successful interaction with surroundings.

While for 2D perception tasks mainly cameras are used, there is a variety of sensors that can be used for 3D perception tasks, ranging from inexpensive solutions like single or multiple cameras to more costly ones like Lidar or Radar.
Lidar is currently a well-adopted choice for 3D perception methods as it creates a set of 3D points forming a point cloud taken by shooting laser beams in multiple directions and counting the time needed for a reflection to be sensed. Point cloud data, despite being sparse and irregular, contains much more information needed for 3D perception compared to images, and is more robust to changes in weather and lightning conditions. 3D object detection and tracking methods using point clouds provide the best combination of accuracy and inference speed, as can be seen in the KITTI leaderboard \cite{kitti_leaderboard}.

Lidars usually operate at 10–20 FPS, and therefore, a method receiving point cloud data as input can be called real-time if its inference speed is at the rate of the Lidar's data generation, as it will not be able to receive more data to process.
Even when the method cannot benefit from the FPS, higher than the Lidar's frame rate, perception methods are usually paired with another method that makes use of the perception results, as in planning tasks \cite{badue2021selfdrivingcars}.
This means that the saved processing time in between Lidar's frames can be used to analyze the results of the tracking without affecting its performance.
In the case of insufficient total frame rate, that can happen due to a slower computing system and higher additional computational load, tracking methods can suffer from dropped frames, that cannot be processed because the system is busy processing a previous data frame. This will result in worse tracking results and, in the case of a high drop rate, it may make the method unusable.

3D SOT, in contrast to the multiple 3D object tracking (3D MOT) task, focuses on tracking a single object of interest with a given initial frame position. This task lies between object detection and multiple object tracking tasks, as the latter one requires objects to be detected first, and then to associate each of them to a previously observed object.
SOT methods do not rely on object detection, but they try to find the object offset on a new frame either by using correlation filters \cite{bolme2010tracking_filters, henriques2015kcf}, or deep learning models that regress to the predicted object offset \cite{held2016goturn}, or use a Siamese approach to find the position with the highest similarity \cite{fang20213dsiamrpn,  bertinetto2016siamfc, li2018siamrpn, li2018siamrpnpp}, or use voting \cite{qi2020p2b, shan2023ptt}.

Application of tracking methods for real-world tasks meets a problem where, due to the model's latency, not all data frames can be processed \cite{li2020towards, li2021dl_for_AD}.
For single-frame tasks, such as object detection, this problem only influences the latency of the predictions and not their quality, but in tracking, the connection between consecutive frames is important. This means that dropping frames can result in wrong associations for MOT or in a lost object for SOT.
Methods like those mentioned above do not take into consideration the limited computational capabilities of embedded computing devices, which are typically used in robotics applications, e.g., self-driving cars, due to a lower power consumption that allows the robotic system to stay active for a longer time.
These devices have a different architecture of CPU-GPU communication than desktop computers and high-end workstations. Thus, depending on the CPU-GPU workload balance and communication protocol of a robotic system, some methods can be better suited for achieving real-time operation. 

In this paper, we present a novel method for 3D SOT receiving voxel pseudo images as an input, as opposed to point-based models proposed for 3D SOT. Voxel pseudo images are created in Bird's-eye View (BEV) coordinates, and since the objects in such projection do not change size, we remove the standard multiscale search process followed by 2D SOT methods, and instead we propose the use of multi-rotation search, where differently rotated search regions are compared against a single target representation to predict both the position and rotation of the object. We show that adopting such an approach leads to fast tracking with competitive performance compared to the point-based methods.
Additionally, the use of voxel pseudo images opens a possibility to adapt 2D SOT methods for a 3D task with minor changes since the input data, structured with pillars, can be directly processed by 2D CNNs.
Considering real-world limitations, we follow \cite{li2020towards} and implement a benchmark for real-time Lidar-based 3D SOT on embedded devices, showcasing the performance drop of the fastest methods when implemented in a real-world scenario.
We compare VPIT using the real-time benchmark with some of the fastest methods, P2P \cite{qi2020p2b} and PTT \cite{shan2023ptt} and show that the performance of those methods degrades significantly when they need to run on embedded devices. This is also an indication that other methods, which are $1.5$ times, or more, slower than these, will not be able to track at all.

Our main contributions can be summarized as follows:
\begin{itemize}
    \item We propose a method that is designed to utilize the specifics of the architecture of embedded devices, which allows transferring data between the GPU and the CPU instantaneously. The proposed method achieves the highest inference speed among the state-of-the-art and loses less speed when applied to embedded devices.
    \item We implement experiments with real-time constraints on data acquisition, which show that while other methods fail to track under these constraints on embedded devices, VPIT has small or even no losses in performance, depending on the devices used.
    \item The architecture of VPIT allows for serving as a bridge between 2D and 3D Siamese SOT methods by showing how an effective 2D Siamese tracking method can be adapted to be used for 3D SOT. This means that future better-performing 2D SOT methods can be easily adapted for being used in 3D SOT based on VPIT.
\end{itemize}

The remainder of the paper is organized as follows: Section \ref{sec:related-works} provides a description of the related works. Section \ref{sec:method} describes the proposed method, along with the proposed training and inference processes. Section \ref{sec:experiments} provides evaluation results on high-end and embedded devices with consideration of real-time requirements. Section \ref{sec:conclusions} concludes the paper and formulates directions for future work.

\section{Related works}\label{sec:related-works}
SOT in 3D is commonly formulated as an extension of 2D SOT. In both cases, an initial position of the object of interest is given, and the method needs to predict the position of the object in all future frames. The main difference between 2D and 3D SOT rises from the type of data used as input. Camera-based 3D perception methods commonly achieve poor performance due to the increased difficulty of extracting correct spatial information from cameras, and therefore most of the 3D object tracking methods use point cloud data or a combination of point clouds and camera images. Point cloud sparsity does not allow using straightforward extensions of 2D SOT methods based on regular CNNs.

SC3D \cite{giancola2019sc3d} uses a Siamese approach by encoding a target point cloud shape in the initial frame and searching for regions in a new frame with the smallest cosine distance between target and search encodings. After finding a new location, the new object shape is combined with the previous one to increase the quality of comparison in future frames. It achieves good tracking results, but its operation is computationally expensive.
Assuming that an object should not move far away from its current location in consecutive frames, one can define a search region inside its neighborhood for searching it in a new frame. This is commonly done by expanding the region around the position of the target.

P2B \cite{qi2020p2b} uses a point-wise network to process points in target and search regions to create a similarity map and find potential target centers. These regions are then used by a voting algorithm to find the best position candidate. 
BAT \cite{zheng2021box} is a 3D tracking method based on P2B, which uses Box Cloud representations as point features, i.e., a representation which depicts the distances between the points of an object and the center and corners of its 3D bounding box.

3D-SiamRPN \cite{fang20213dsiamrpn} uses a Siamese point-wise network to create features for target and search point clouds. It then uses a cross-correlation algorithm to find points of the target in a new frame.
An additional region proposal subnetwork is used to regress the final bounding box. 
The F-siamese tracker \cite{zou2020f} aims to fuse RGB and point cloud information for 3D tracking and applies a 2D Siamese model to generate 2D proposals from an RGB image, which are then used for 3D frustum generation. The proposed frustums are processed with a 3D Siamese model to get the 3D object position. 
Point-Track-Transformer (PTT) \cite{shan2023ptt} creates a Transformer module for point-based SOT methods and employs it based on a P2B model, leading to an increased performance.
Siamese Transformer Network (STNet) \cite{hui2022stnet} is a 3D Siamese method that uses a point Transformer network to learn shape context information of an object and proposes an iterative coarse-to-fine correlation network to improve robustness of cross-correlation.
Point Tracking TRansformer (PTTR) \cite{zhou2022pttr} is a Siamese method which uses a modified version of PointNet++ \cite{qi2017pointnetpp} with a proposed Relation-Aware Sampling mechanism to select better points from the search area. Feature matching in PTTR is performed by a proposed Point Relation Transformer with a Relation Attention Module, and the final predictions are done in a coarse-to-fine manner employing a Prediction Refinement Module.
Global-Local Transformer Voting (GLT-T) \cite{nie2023glt} is a Siamese method that focuses on overcoming the drawbacks of VoteNet \cite{qi2019votenet} by proposing a Global-Local Transformer module that incorporates global and local knowledge via self-attention to encode object-aware and patch-aware prior into features of seed points.

3D Siam-2D \cite{zarzar2020efficient} uses two Siamese networks, one that creates fast 2D proposals in BEV space, and another one that uses projected 2D proposals to identify which of the proposals belong to the object of interest.
The loss of information that occurs in BEV projection affects the performance of the 2D Siamese network. Although voxel pseudo images lie on the BEV space, they do not have this problem, as each pixel of this image incorporates information about the points in a corresponding voxel via a small neural network, and not the projection alone.

Recent works on 3D SOT focus on improving speed as well as tracking performance, but most do not take into consideration the limited computational capabilities of embedded devices, like the NVIDIA Jetson series, which are commonly used in robotics applications.
Large delays in processing incoming frames can lead to dropped frames, which can quickly lead to performance degradation of traditional trackers as larger and larger offsets must be predicted as time progresses. In 2D tracking, this concept has been investigated, for example in the real-time experiments of the VOT benchmarks \cite{kristan2021ninth}, and more recently by \cite{li2020towards} for streaming perception tasks in general.
In this work, we extend this notion to 3D SOT, and design VPIT to be used efficiently and effectively on embedded devices.
Furthermore, to the best of our knowledge, VPIT is the first tracker to use a siamese architecture on point cloud pseudo images directly, while using varying rotations of the target area to find the target's rotation in subsequent frames. VPIT can be seen as a 3D extension of 2D siamese-based trackers \cite{bertinetto2016siamfc,nousi2020dense}, paving the way for other 2D approaches to be successfully extended to the 3D case, while maintaining high tracking speed, even on embedded devices.

\section{Proposed method}\label{sec:method}
Our proposed method is based on a modified PointPillars architecture for 3D single object tracking, which is originally trained for 3D object detection.
In the following subsections, we first introduce our proposed modifications to the baseline architecture to shift from detection to tracking, and then describe the training and inference processes. Implementation of our method is publicly available in the OpenDR toolkit\footnote{\url{https://github.com/opendr-eu/opendr}}.

\subsection{Model architecture}\label{sec:method-architecture}

Point cloud data is irregular and cannot be processed with 2D convolutional neural networks directly. This forces methods to either structure the data using techniques such as voxelization \cite{2017voxelnet, lang2018pointpillars, liu2019tanet, 2019hotspotnet}, or to use neural networks that work well on unordered data, such as MLPs with maximum pooling operations or Transformers \cite{qi2017pointnet, qi2017pointnetpp, fang20213dsiamrpn, qi2020p2b, shan2023ptt}.

Existing datasets for 3D object detection and tracking, such as KITTI \cite{2012kitti} and NuScenes \cite{nuScenes2019}, consider only a single rotation angle: around the vertical axis. This limitation raises from the nature of scenes in these datasets, as they present data for outdoor object detection and tracking with 
objects moving on roads. 
These objects are mostly rotated in Bird's-eye View (BEV) space, and therefore only this rotation is considered for simplicity.

Since the rotation is limited to the BEV space, we can perform 2D-like Siamese tracking on a structured point cloud representation, such as PointPillars pseudo image \cite{lang2018pointpillars}.
This image is the result of voxelization, where the vertical size of each voxel takes up the whole vertical space, and therefore creating a 2D map in BEV space where each pixel represents a small subspace of a scene and points inside it.
Objects in such an image have constant size, as they are not affected by projection size distortions, but their rotation makes the Axis-Aligned Bounding Boxes (AABB) tracking impractical.

Siamese models use an identical transformation $\theta(\cdot)$ to both inputs $x$ and $z$, which are then combined by some function $g(\cdot)$, i.e. $f(x, z)=g(\theta(x), \theta(z))$. 
Siamese tracking methods select $\theta(\cdot)$ to be an embedding function and $g(\cdot)$ to be a similarity measure function.
If the input $x$ is considered to be a target image that we want to find inside the search image $z$, the output of such a model is a similarity score map, that has high values on the most probable target object locations.

Tracking is performed by first initializing the target region $t_0$ with the given object location and a corresponding search region $s_0 = \sigma(t_0)$ that should be big enough to accommodate for possible object position offset during the time difference between the input frames.
For the frames $F_{\tau - 1}$ and $F_{\tau}$, the target region from the previous frame $t_{\tau-1}$ and the corresponding search region $s_{\tau - 1}$ are processed by applying the Siamese model to create the next target and search regions:
\begin{align}
\begin{split}
    t_{\tau} &= t_{\tau - 1} + \delta(f(\xi_t(t_{\tau - 1}), \xi_s(s_{\tau - 1}))), \\
    s_{\tau} &= \sigma(t_{\tau}),
\end{split}
\end{align}
where the $\delta(\cdot)$ function transforms the similarity map into the target position offset and $\xi_t(\cdot)$ and $\xi_s(\cdot)$ functions transform the target and search regions, respectively, into an image which is processed by the embedding function $\theta(\cdot)$.
More details about these functions in Section \ref{sec:method-training}.
This process is repeated for each new frame to update the object location.

SiamFC \cite{bertinetto2016siamfc} is a lightweight 2D SOT method that achieves the highest inference speed among 2D SOT methods. It uses a fully convolutional network as $\theta(\cdot)$ and creates a set of search regions for each target to address the possible change in size due to the projection distortion.
We follow the approach of SiamFC and use PointPillars \cite{lang2018pointpillars} to structure point clouds by generating voxel pseudo image, which serves as an input to the Siamese model, that uses a reduced PointPillars' Region Proposal Network (RPN) as $\theta(\cdot)$ and a similarity function $g(a,b) = a \otimes b$, where $\otimes$ indicates the correlation operator. In practice, because most DL frameworks perform correlation in convolutional layers, the similarity function can be implemented as $g(a,b) = \mathrm{conv2D}_{\omega=b}(a)$, where $\omega$ are the weights of the layer.
\begin{figure}[!ht]
\begin{center}
    \includegraphics[width=0.95\linewidth]{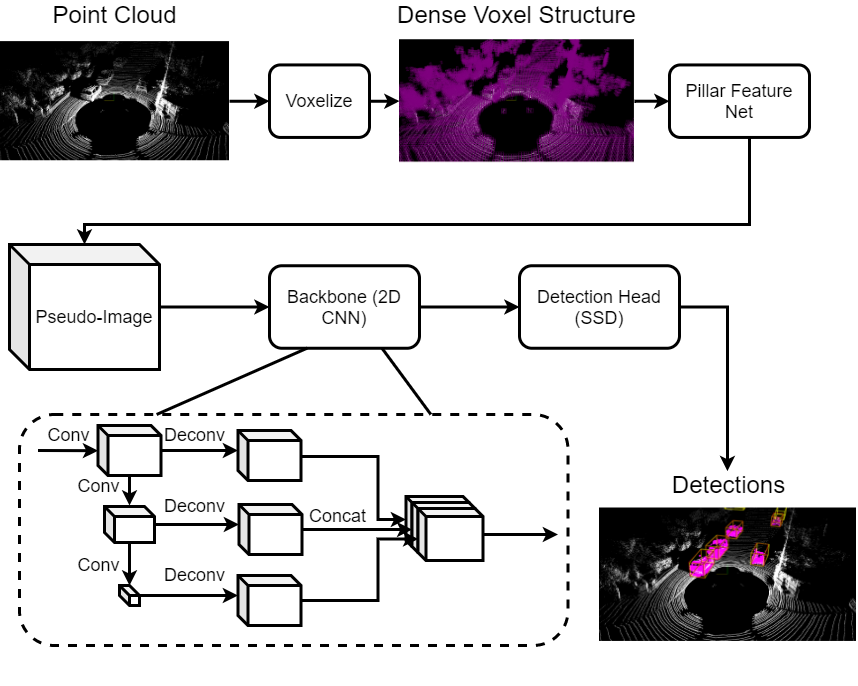}
    \caption{Structure of PointPillars 3D object detection model. The RPN is a 2D CNN that takes a pseudo image as input.}
    \label{fig:pointpillars}
\end{center}
\end{figure}

\begin{figure*}[!t]
\begin{center}
    \includegraphics[width=1\linewidth]{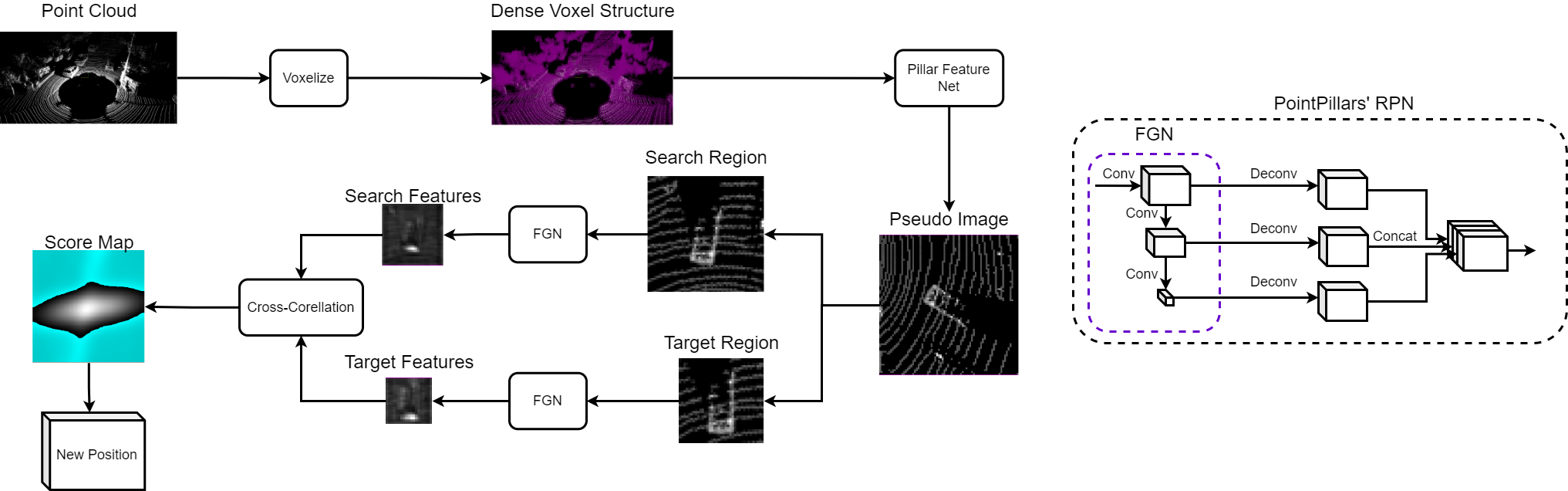}
    \caption{Structure of the proposed Voxel Pseudo Image Tracking model. The input point cloud is voxelized and processed with the PointPillars' Pillar Feature Network to create a voxel pseudo image, which serves as an input to the Siamese model. The Feature Generation Network (FGN), which is a convolutional subnetwork of the PointPillars' RPN, processes the target and search regions to create corresponding features that are then compared to find a position of the best similarity.}
    \label{fig:model}
\end{center}
\end{figure*}

The Region Proposal Network (RPN) in PointPillars is responsible for processing the input pseudo image using a fully convolutional network with 3 blocks of convolutional layers and 3 transposed convolutions that create same-sized features for final box regression and classification, as can be seen in Fig. \ref{fig:pointpillars}.
For Siamese tracking purposes, we are only interested in feature generation and there is no need for box regression and classification parts of the RPN. Therefore, we select convolutional blocks of the RPN to be used as a Feature Generation Network (FGN) $\theta(\cdot)$.
The architecture of the model is shown in Fig. \ref{fig:model}.
The initial pipeline is similar to PointPillars for Detection, which includes voxelization of the input point cloud and processing it with the Pillar Feature Network to create a voxel pseudo image.
The target and search regions of the pseudo image are processed by a Feature Generation Network, creating features that are compared by a cross-correlation module to find a position of the highest similarity, which serves as a new target position.

Each of the target ($t$) and search ($s$) regions are represented by a set of 5 values $(x,y,w,h,\alpha)$, where $(x, y)$ is the position of the region center in pseudo image space in pixels, $(w, h)$ is the size of the region and $\alpha$ is the rotation angle.
The ground truth bounding box $B_{gt}$ is described by the 3D position $(x, y, z)$, size $(w, h, d)$ and rotation $\alpha$.
The corresponding target and search regions are created as follows:
\begin{align}
\begin{split}
    t_{0} &= \kappa_c((B^x_{gt}, B^y_{gt}, B^w_{gt}, B^h_{gt}, B^{\alpha}_{gt})), \\
    s_{0} &= \sigma(t_{0}),
\end{split}
\end{align}
where $t_{0}$ is the initial target region, $s_{0}$ is the initial search region and the $\kappa_c(\cdot)$ is a function that adds context to the target region based on the amount of context parameter $c$:
\begin{align}
\begin{split}
    &\kappa_{c}(t) =
    \begin{cases}
        \kappa_{c_+}(t), & \text{if } c > 0, \\
        \kappa_{c_-}(t), & \text{otherwise},
    \end{cases} \\
    &\kappa_{c_+}(x,y,w,h,\alpha) = (x, y, m_n, m_n, \alpha), \\ 
    &m_n = \sqrt{(w+m)(h+m)}, \\
    &m = c (w + h), \\
    &\kappa_{c_-}((x,y,w,h,\alpha)) = (x, y, w(1-c), h(1-c), \alpha), \\ 
\end{split}
\end{align}
where $\kappa_{c_+}(\cdot)$ adds context to the object by making a square region that includes the original region inside, and $\kappa_{c_-}(\cdot)$ adds context by increasing each side of the region independently.
The $\sigma(\cdot)$ function creates a search region, the size of which is defined by a hyperparameter $\sigma_s$ indicating the search region to target region size ratio, i.e.:
\begin{align}
\begin{split}
    &\sigma(x,y,w,h,\alpha) = (x,y,\sigma_s w,\sigma_s h,\alpha). \\
\end{split}
\end{align}
The predicted output for the frame $\tau$ is computed as:
\begin{equation}
    B_{\tau} = (t^x_{\tau}, t^y_{\tau}, B^z_{gt}, t^w_{\tau}, t^h_{\tau}, B^d_{gt}, t^{\alpha}_{\tau}),
\end{equation}
where $B^z_{gt}$ and $B^d_{gt}$ are $z$ and depth values, respectively, of the provided initial bounding box.

\subsection{Training}\label{sec:method-training}
We start from a pretrained PointPillars model on KITTI Detection dataset.
The training can be performed on both KITTI Detection and KITTI Tracking datasets.
Training on KITTI Detection dataset is done by considering objects separately and creating target and search region from their bounding boxes.
For a set of ground truth boxes $\{B_{\mu_i} \: | \: i \in [1, N]\}$ in an input frame $\mu$, where $N$ is a number of ground truth objects in this frame, we create $N$ training samples by considering target-search pairs $\{(t_{\mu_i}, s_{\mu_i}) \: | \: i \in [1, N]\}$ created from these ground truth boxes as follows:
\begin{align}
\begin{split}
    t_{\mu_i} &= \kappa_c((B^x_{\mu_i}, B^y_{\mu_i}, B^w_{\mu_i}, B^h_{\mu_i}, B^{\alpha}_{\mu_i})), \\
    s_{\mu_i} &= \sigma_a(\sigma(t_{\mu_i}), t_{\mu_i}),
\end{split}
\end{align}
where $\kappa_c(\cdot)$ and $\sigma(\cdot)$ functions are identical to the ones, described before, and $\sigma_a(\cdot)$ represents an augmentation technique where the center of the search region is shifted from the target center to imitate object movement between frames:
\begin{equation}
    \sigma_a(s, t) = (t^x + \epsilon_x, t^y + \epsilon_y, s^w, s^h, s^\alpha),
\end{equation}
with $\epsilon_x \sim \mathrm{uniform}(-\frac{s^w - t^w}{2}, \frac{s^w - t^w}{2})$ and $\epsilon_y \sim \mathrm{uniform}(-\frac{s^h - t^h}{2}, \frac{s^h - t^h}{2})$.
In addition to the proposed augmentation, we use point cloud and ground truth boxes augmentations used in PointPillars, which include point cloud translation, rotation, point and ground truth database sampling.

For training on KITTI tracking dataset, we select the target and search regions from the same track $k$ and object $o_k$, but the corresponding point clouds and bounding boxes are taken from the different frames, modeling the variance of target object representation in time:
\begin{align}
\begin{split}
    t_{o_k} &= \kappa_c((B^x_{f_t}, B^y_{f_t}, B^w_{f_t}, B^h_{f_t}, B^{\alpha}_{f_t})), \\
    \hat{t}_{o_k} &= \kappa_c((B^x_{f_s}, B^y_{f_s}, B^w_{f_s}, B^h_{f_s}, B^{\alpha}_{f_s})), \\
    s_{o_k} &= \sigma_a(\sigma(\hat{t}_{o_k}), \hat{t}_{o_k}),
\end{split}
\end{align}
where $f_t$ and $f_s$ are a randomly selected frames from the track $k$ that contain the object of interest $o_k$.
We balance the number of occurrences of the same object by selecting a constant number of $(f_t, f_s)$ samples for each object in the dataset.

After creating target and search regions, we apply $\xi_t(\cdot)$ and $\xi_s(\cdot)$ functions by first taking sub-images of those regions and then, depending on the interpolation size parameter, applying bicubic interpolation to have a fixed size of the image:
\begin{align}
\begin{split}
    &\nu(x) = \mathrm{subimage}(\Pi(x), x), \\
    &\xi_\rho(x) = 
    \begin{cases}
        \mathrm{bicubic}(\nu(x),\iota_\rho), & \text{if } \iota_\rho > 0, \\
        \nu(x), & \text{otherwise},
    \end{cases}
\end{split}
\end{align}
where $x$ is a region to be processed, $\rho$ indicates that either target region $t$, or search region $s$ is selected, $\iota_\rho$ is an interpolation size parameter, which indicates the output size of the bicubic interpolation, and $\Pi(x)$ creates an AABB pseudo image from the input point cloud that contains the region $x$.
These images are processed with the Feature Generation Network, which is the convolutional subnetwork of the PointPillars' RPN, as shown in Fig. \ref{fig:model}.

The resulting features are compared using the cross-correlation module to create a score map.
A Binary Cross-Entropy (BCE) loss is used to compare a predicted score map $M_x$ to the true label $M_y$ as follows:
\begin{align}
\begin{split}
    &BCE(M_x, M_y) = \frac{1}{N} \sum_1^N l_n, \\
    &l_n = -w_n \Big(M_{y_n} \log S(M_{x_n}) + (1-M_{y_n}) \log (1-S(M_{x_n})) \Big),
\end{split}
\end{align}
where $S(x) = 1 / (1 + e^{-x})$ and $w_n$ is a scaling weight for each element.
We follow SiamFC \cite{bertinetto2016siamfc} and select the values of $w_n$ to equalize the total weight of positive and negative pixels on the ground truth score map.

The true label is created by placing positive values on a score map within a small distance of the projected target center, and zeros everywhere else. The value $v(p_x,p_y)$ of a pixel $(p_x,p_y)$ in a label map depends on its distance to the target center $d(p_x,p_y)$ and a hyperparameter $r$, which describes the radius of positive labels around the center:
\begin{align}
\begin{split}
    &v(x,y) = \\
    &\begin{cases}
        v_{\mathrm{min}} \frac{d(p_x,p_y)}{r} + v_{\mathrm{max}} (1 - \frac{d(p_x,p_y)}{r}), & \text{if } d(p_x,p_y) \leq r + 1, \\
        0, & \text{otherwise},
    \end{cases}\\
    &d(p_x,p_y) = \sqrt{(p_x-c_x)^2 + (p_y-c_y)^2},
\end{split}
\end{align}
where $(c_x, c_y)$ is the target center, $[v_{\mathrm{min}}, v_{\mathrm{max}}]$ is the range of values assigned to the pixels inside the circle of radius $r$, and the pixels inside the $(r, r+1]$ range are assigned to values, lower than $v_{\mathrm{min}}$.

\subsection{Inference}\label{sec:method-inference}
Inference is split into two parts, i.e., initialization and tracking.
The 3D bounding box of the object of interest is given by either a detection method, a user, or from a dataset, to create the initial target and its features, that will be used for the future tracking. 
During tracking, the last target position is used to place a search region, but instead of using only a single search region, we create multiple search regions with identical size and different rotations.
This is done to allow to correct for rotation changes in the object. 
The set of $2K + 1$ search regions $\{s_{\tau_i} \: | \: i \in [-K, K]\}$ is created by altering the initial search region $s_{\tau_0}$, i.e., $s^{x,y,w,h}_{\tau_i} = s^{x,y,w,h}_{\tau_0}$ and $s^\alpha_{\tau_i} = s^\alpha_{\tau_0} + i \psi_\alpha$, where $\psi_\alpha$ is a rotation step between consequent search regions.

Target features are compared with all search features, and the one with the highest score is selected for a new rotation:
\begin{align}
\begin{split}
    &s_{\tau_{\mathrm{max}}} = \underset{s_{\tau_i}}{\mathrm{argmax}} \:\: \Lambda_i \max (f(\xi_t(t_{\tau}), \xi_s(s_{\tau_i}))), \\
    &t^\alpha_{\tau + 1} = \mu s^\alpha_{\tau_{\mathrm{max}}} + (1 - \mu) t^\alpha_{\tau},
\end{split}
\end{align}
where $\Lambda_i$ is a rotation penalty multiplier, which is $1$ for $i=0$ and a hyperparameter value in $[0,1]$ range for all other search regions, $\mu$ is a rotation interpolation coefficient and $f$ is a Siamese model.

The score map $M$, obtained from the Siamese model, is upscaled with bicubic interpolation, increasing its size $16$ times by default (score upscale parameter $u_M$).
Based on the assumption that an object cannot move far from its previous position in consecutive frames, we use a penalty technique for the scores that are far from the center, by multiplying scores with a weighted penalty map.
The penalty map is formed using either Hann window or a 2D Gaussian function.
The maximum score position $(x, y)_{\mathrm{max}}$ of the upscaled score map is translated back from the score coordinates to image coordinates as follows:
\begin{align}
\begin{split} \label{eq:scoremap}
    &(x, y)_{\mathrm{max}} = \underset{(x, y)}{\mathrm{argmax}} \:\: H(f(\xi_t(t_{\tau}), \xi_s(s_{\tau_{\mathrm{max}}})))_{(x, y)}, \\
    &(x, y)^{\mathrm{img}}_{\mathrm{max}} = (x, y)_{\mathrm{max}} \frac{s^{(w, h)}_{\tau_{\mathrm{max}}}}{H_{\mathrm{size}}}  , \\
    &H_{\mathrm{size}} = u_M M_{\mathrm{size}}, \\
    &H(M) = \eta P(H_{\mathrm{size}}) + (1 - \eta) \: \mathrm{bicubic}(M, u_M, M_{\mathrm{size}}),
\end{split}
\end{align}
where $(x, y)^{\mathrm{img}}_{\mathrm{max}}$ is a target position offset, which represents a prediction of the target object's movement, $H$ applies score interpolation and a penalty map $P(\mathrm{size})$ to the upscaled output of the Siamese model, $M_{\mathrm{size}}$ and $H_{\mathrm{size}}$ are the 2D sizes of the original score map and the upscaled score map, respectively, and $\eta$ is a window influence parameter that controls how much the Hann window or a Gaussian penalty influence the score map.

After the current frame's prediction is ready, the search region is centered on a prediction, assuming that the object's position in the next frame should be close to its position in the previous frame.
In order to make it easier to find the position of an object inside a search region, we apply a linear position extrapolation for the search region by centering it not on a previous prediction, but on a possible new prediction position.
Given a target region from the last frame $t_{\tau-1}$ and the predicted target region $t_\tau$, the corresponding search region has its position defined $s_\tau^{x,y} = 2 t_\tau^{x,y} - t_{\tau-1}^{x,y}$. 
This approach increases the chances of the target to be in the center of the search region or having a small deviation from it.

\begin{figure}
     \centering
     \begin{subfigure}[b]{0.2\linewidth}
         \centering
         \includegraphics[width=\linewidth]{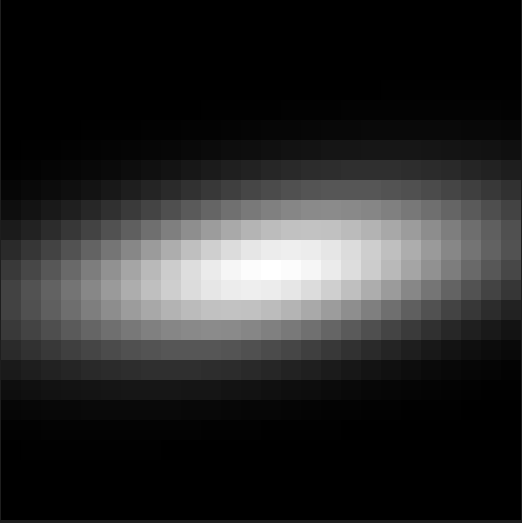}
         \caption{Gaussian penalty}
     \end{subfigure}
     \hfill
     \begin{subfigure}[b]{0.2\linewidth}
         \centering
         \includegraphics[width=\linewidth]{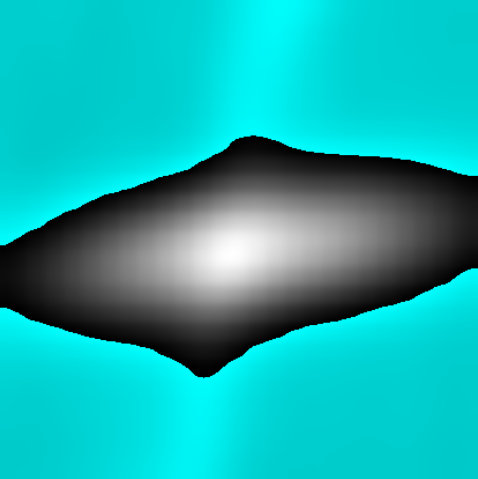}
         \caption{Estimated score map}
     \end{subfigure}
     \hfill
     \begin{subfigure}[b]{0.2\linewidth}
         \centering
         \includegraphics[width=\linewidth]{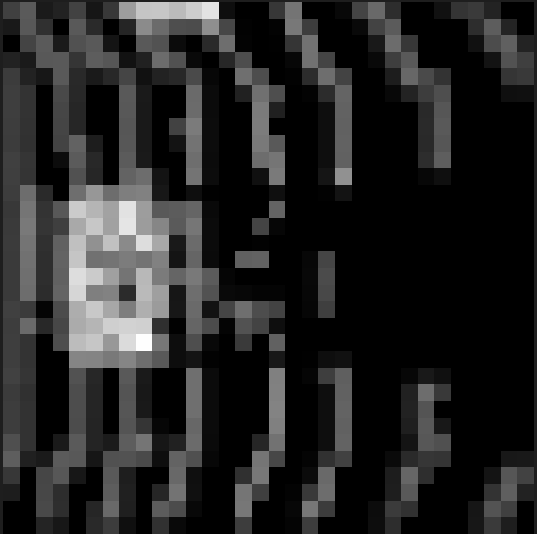}
         \caption{Target pseudo image}
     \end{subfigure}
     \hfill
     \begin{subfigure}[b]{0.2\linewidth}
         \centering
         \includegraphics[width=\linewidth]{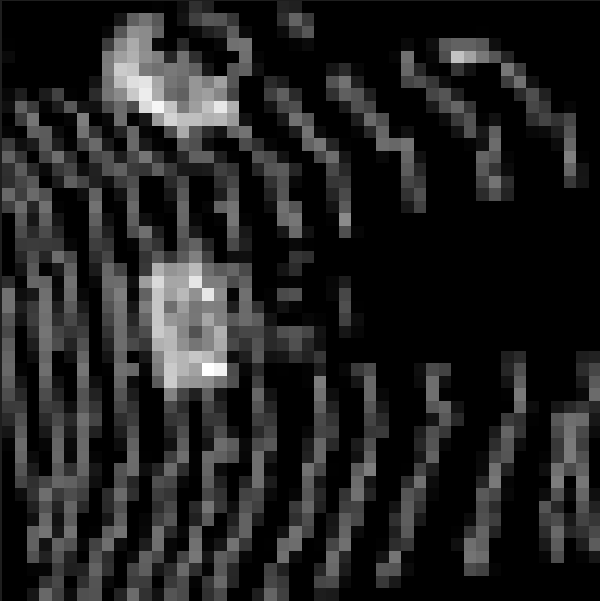}
         \caption{Search pseudo image}\label{fig:penalty_and_regions:d}
     \end{subfigure}
        \caption{An example of a directional Gaussian penalty used with linear extrapolation, a corresponding score map and target and search pseudo images. The cyan color on the score map represents negative values.}
        \label{fig:penalty_and_regions}
\end{figure}

When the linear extrapolation is used, the penalty map $P(\mathrm{size})$ (Eq. \ref{eq:scoremap}) is created using a 2D Gaussian with a covariance matrix $\Sigma$ that represents the angle of the extrapolation vector to penalize more positions that are not on the way of the predicted object's movement.
This is done by creating a Gaussian with independent variables and variances $\sigma_+$ for the desired direction and $\sigma_-$ for the opposite direction:
\begin{equation}
    \mu = 0 \:\:\:\:\:\:\:\: \textrm{and} \:\:\:\:\:\:\Sigma_0 =
    \begin{bmatrix}
        \sigma_- & 0 \\
        0 & \sigma_+
    \end{bmatrix}.\label{eq:extrapolation_sigma0}
\end{equation}
Bigger $\sigma_+$ allows a further offset alongside the extrapolation vector, while bigger $\sigma_-$ allows a higher deviation from it. 
The resulting Gaussian $\mathcal{N}(\mu, \Sigma_0)$ is rotated by $\phi = \arctan(e)$, where $e$ is the extrapolation vector, as follows:
\begin{equation}
    R = \begin{bmatrix}
        \cos(\phi) & -\sin(\phi) \\
        \sin(\phi) & \cos(\phi)
    \end{bmatrix} \:\:\:\:\:\:\:\: \textrm{and} \:\:\:\:\:\:\ \Sigma = R \Sigma_0^{-1} R^T.\label{eq:extrapolation_rotation}
\end{equation}
The creation of such a map for a high-dimensional upscaled score map is a computationally expensive task, and therefore, to optimize it, the penalty maps are hashed by their size and rotation to reuse in future frames.
There is no need in keeping high precision of rotation value, as the penalty map will be almost identical for close $\phi$ values and will result in the same basis for predictions.
Keeping this in mind, we define a rotation hash function ~$H_r(\phi) = \left \lfloor \frac{n\phi}{2\pi} \right \rfloor$. 
This hash function divides the full circle on $n$ equal regions, and the penalty map is taken for the closest $\phi$-sector.
Fig. \ref{fig:penalty_and_regions} shows an example of a Gaussian penalty created using this approach, a corresponding score map and target and search pseudo images.

\begin{figure}
     \centering
     \begin{subfigure}[b]{0.2\linewidth}
         \centering
         \includegraphics[width=\linewidth]{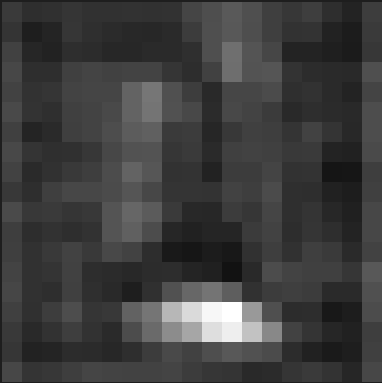}
         \caption{$\tau=0$}
     \end{subfigure}
     \hfill
     \begin{subfigure}[b]{0.2\linewidth}
         \centering
         \includegraphics[width=\linewidth]{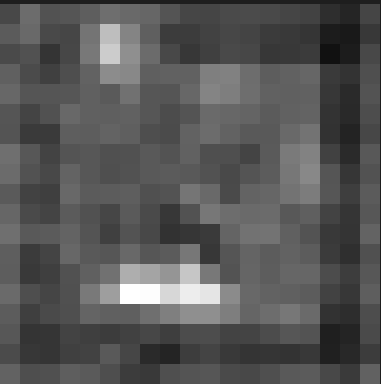}
         \caption{$\tau=40$}
     \end{subfigure}
     \hfill
     \begin{subfigure}[b]{0.2\linewidth}
         \centering
         \includegraphics[width=\linewidth]{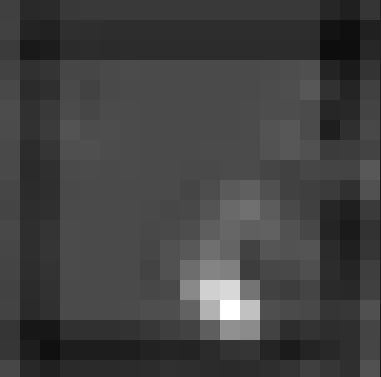}
         \caption{$\tau=60$}
     \end{subfigure}
     \hfill
     \begin{subfigure}[b]{0.2\linewidth}
         \centering
         \includegraphics[width=\linewidth]{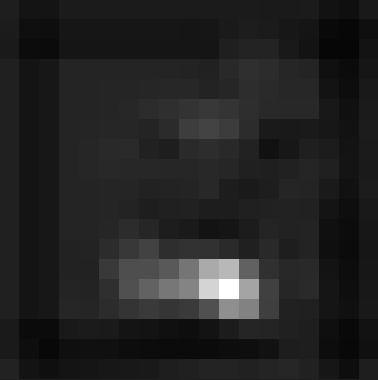}
         \caption{$\tau=80$}
     \end{subfigure}
        \caption{Target features of the same object at different frames.}
        \label{fig:target_features}
\end{figure}

\begin{figure}
     \centering
     \begin{subfigure}[b]{0.2\linewidth}
         \centering
         \includegraphics[width=\linewidth]{target_feature_0.png}
         \caption{$\tau=0$}
     \end{subfigure}
     \hfill
     \begin{subfigure}[b]{0.2\linewidth}
         \centering
         \includegraphics[width=\linewidth]{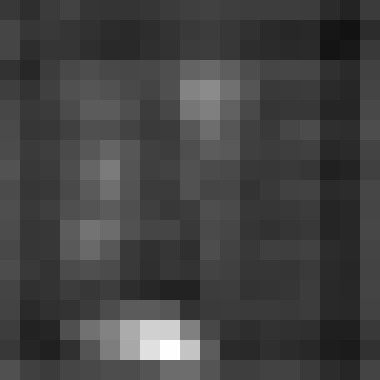}
         \caption{$\tau=10$}
     \end{subfigure}
     \hfill
     \begin{subfigure}[b]{0.2\linewidth}
         \centering
         \includegraphics[width=\linewidth]{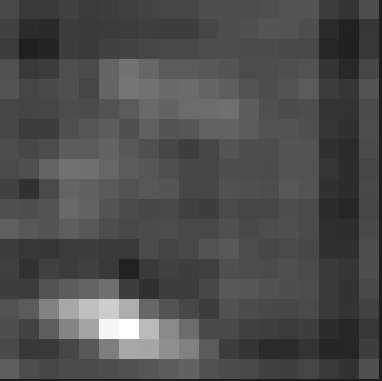}
         \caption{$\tau=20$}
     \end{subfigure}
     \hfill
     \begin{subfigure}[b]{0.2\linewidth}
         \centering
         \includegraphics[width=\linewidth]{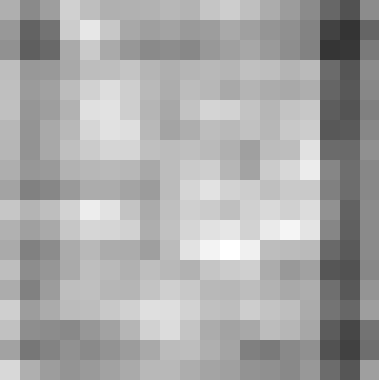}
         \caption{$\tau=30$}
     \end{subfigure}
        \caption{Merged target features of the same object at different frames with $0.5$ merge scale. Starting from frame 10, the object representation drifts to the right and completely loses the target at frame 30.}
        \label{fig:target_features_05}
\end{figure}

Target features are created at the first frame and then used to find the object during the whole tracking sequence.
However, due to a high variance of object representation at different distances to Lidar, initial features may be too far from the current representation of the object, as can be seen from Fig. \ref{fig:target_features}.
This problem may be resolved by using target features from the latest frame, but such an approach leads to an error accumulation and drifts the target region to a background object.
In order to balance between the aforementioned approaches, we introduce a target feature merge scale $m_{\mathrm{tf}}$.
Using the $m_{\mathrm{tf}} = 0$, only the initial target features will be used through the tracking task, while using the $m_{\mathrm{tf}} = 1$ results in the latest frame's target features overwriting the previous ones.
Fig. \ref{fig:target_features_05} shows the effect of a high merge scale value $m_{\mathrm{tf}} = 0.5$.
The target region is slowly drifting to the right starting from frame 10 and the object is completely lost at frame 30.
We use small values of $m_{\mathrm{tf}}$ in range $[0, 0.01]$ in ablation study.
This allows to keep the target features stable and not drift out from the object, while still having the possibility to update features if the distance to the object changes. 

By default, the model's position offset predictions are applied directly, but we can improve prediction stability by introducing the offset interpolation parameter $\omega$.
Given the previous frame's target position $\zeta_{\tau - 1}$ and a prediction for the current frame $\hat{\zeta}_{\tau}$, the interpolated position $\zeta_{\tau}$ is defined as follows:
\begin{equation}
    \zeta_{\tau} = \omega \zeta_{\tau - 1} + (1 - \omega) \hat{\zeta}_{\tau}.
\end{equation}

Current 3D detection and tracking datasets provide annotations for a rotation around the vertical axis only, which is sufficient for current robotic and autonomous driving tasks, but it may be too coarse for future tasks.
For datasets with rotations around all axes and not only the vertical one, an additional regression branch can be added that, applied to a target region, will predict other rotation angles.
A similar process can be applied to non-rigid objects in order to continuously update their dimensions.

\begin{table*}[]
        \caption{Results of 3D Car tracking on KITTI dataset. Modality represents the type of data the tracking is performed on (PC for point cloud, BEV for Birds-Eye-View and VPI for voxel pseudo image). FPS values are reported on a 1080Ti GPU by a corresponding paper. FPS values with a star notation are obtained by running the methods' official implementations on a 1080ti GPU, considering the full runtime of the network.}
        \label{tab:results-reported}
        \centering
        \begin{tabular}{l|cccc}
            \toprule
            \textbf{Method} & \textbf{Modality} & \textbf{Success} & \textbf{Precision} & \textbf{FPS} \\
            \midrule
            3DSRPN PCW \cite{fang20213dsiamrpn} & PC & 56.32 & 73.40 & 16.7 \\
            3DSRPN PW \cite{fang20213dsiamrpn} & PC & 57.25 & 75.03 & 20.8 \\
            SCD3D-KF \cite{giancola2019sc3d} & PC & 40.09 & 56.17 & 2.2 \\
            SCD3D-EX \cite{giancola2019sc3d} & PC & \textbf{76.94} & 81.38 & 1.8 \\
            3D Siam-2D \cite{zarzar2020efficient} & PC+BEV & 36.3 & 51.0 & - \\
            GLT-T \cite{nie2023glt} & PC & 68.2 & 82.1 & 29.94 \\
            \midrule
            STNet \cite{hui2022stnet} & PC & 70.58 & \textbf{81.91} & 15.73* \\
            BAT \cite{zheng2021box} & PC & 65.38 & 78.88 & 23.96* \\
            P2B \cite{qi2020p2b} & PC & 56.2 & 72.8 & 30.18* \\
            PTTR \cite{zhou2022pttr} & PC & 65.19 & 77.39 & 32.51* \\
            PTT-Net \cite{shan2023ptt} & PC & 67.8 & 81.8 & 39.51* \\
            \midrule
            VPIT (Ours) & VPI & 50.49 & 64.53 & \textbf{50.45}\\
            \bottomrule
        \end{tabular}
\end{table*}

\begin{center}
    \begin{table*}[!ht]
        \caption{Evaluation of some of the fastest methods for 3D Car tracking on KITTI dataset. The evaluation is performed with official implementations on high-end and embedded GPU platforms with different combinations of GPU/CPU. 32C CPU and 104C CPU correspond to 32-core and 104-core CPUs, respectively.}
        \label{tab:results-fps}
        \centering
        \resizebox{\columnwidth}{!}{%
        \begin{tabular}{l|ccc|ccccc}
            \toprule
            \multirow{3}{*}{\textbf{Method}} & \multirow{3}{*}{\textbf{Modality}} & \multirow{3}{*}{\textbf{Success}} & \multirow{3}{*}{\textbf{Precision}} & \multicolumn{5}{c}{\textbf{FPS}} \\
            & & & & 1080Ti & 2080 & 2080Ti & TX2 & Xavier\\
            & & & & 32C CPU & 32C CPU & 104C CPU &  & \\
            \midrule
            P2B \cite{qi2020p2b} & PC & 56.2 & 72.8 & 30.18 & 26.34 & 38.93 & 6.20 & 10.37 \\
            PTT-Net \cite{shan2023ptt} & PC & \textbf{67.8} & \textbf{81.8} & 39.51 & 33.34 & 50.25 & 8.04 & 13.49 \\
            VPIT (Ours) & VPI & 50.49 & 64.53 & \textbf{50.45} & \textbf{52.52} &  \textbf{72.53} & \textbf{14.61} & \textbf{20.55}\\
            \bottomrule
        \end{tabular}%
        }
    \end{table*}
\end{center}

\section{Experiments}\label{sec:experiments}
We use KITTI \cite{2012kitti} Tracking training dataset split to train and test our model, with tracks 0-18 for training and validation and tracks 19-20 for testing (as is common practice \cite{fang20213dsiamrpn, giancola2019sc3d, zarzar2020efficient, qi2020p2b, shan2023ptt}).
The dataset includes challenges of distant objects with sparse point clouds, partial and full temporal object occlusion, side-to-side objects of the same class, variance in speed of both ego object and the object of interest.
We use Precision and Success metrics as defined in One Pass Evaluation \cite{2016precision_success}.
Precision is computed based on the difference between ground-truth and predicted object centers in 3D.
Success is computed based on the 3D Intersection over Union (IoU) between predicted and ground truth 3D bounding boxes.


We use $1$ feature block from the original PointPillars model with $4$ layers in a block. 
The model is trained for $64,000$ steps with BCE loss, $1*10^{-5}$ learning rate and $2$ positive label radius. During inference, rotations count of $3$, rotation step of $0.15$ and rotation penalty of $0.98$ are used, together with $0.85$ penalty map multiplier, score upscale of $8$, target/search size of $(0, 0)$ (original sizes are used), context amount of $0.27$, rotation interpolation of $1$, offset interpolation of $0.3$, target feature merge scale of $0.005$ and linear search position extrapolation.


\subsection{Comparison with state-of-the-art}

In our experiments we focus on the \textit{Car} class of KITTI because this is the most represented class in the dataset and the state-of-the-art 3D SOT methods place their main focus on the tracking of cars.
Evaluation results on 1080Ti GPU are given in Table \ref{tab:results-reported}.
Even though we are interested in application scenarios involving processing on embedded devices, we first measure the performance and processing speed of the proposed method with competing ones on a 1080Ti, since this is the most commonly used GPU when evaluating the performance of the competing methods. We do this, in order to establish a relative speed measure between the methods.
VPIT is the fastest method and achieves competitive Precision and Success.
Experiments in Table \ref{tab:results-reported} are performed in an offline manner, which means that the tested methods have all the time required to process each frame.
However, such an approach does not consider the limitations under which tracking methods should perform when applied to real-life problems. As shown in Section \ref{sec:experiments-real-time}, considering real-time limitations to the evaluation process illustrates how the superior inference speed of VPIT allows it to retain a high tracking accuracy while the other methods lead to much inferior results.
We evaluate official implementations of P2B and PTT, and VPIT, on different high-end and embedded GPU platforms to showcase how the architecture of the device influences the models' FPS.
Embedded platforms include NVIDIA Jetson TX2 with 8GB of shared (V)RAM and 256 Core Pascal GPU, and NVIDIA Jetson AGX Xavier with 16 GB of shared (V)RAM and 512 Core Volta GPU.

When computing FPS, we include all the time the model spends to process the input and create a final result.
This excludes time spent obtaining data and to computing Success and Precision values, but includes pre- and post-processing steps.

As can be seen from Table \ref{tab:results-fps}, in terms of speed, VPIT outperforms P2B by 67\% on 1080Ti GPU, 99\% on 2080 GPU and 86\% on 2080Ti GPU with 104-core CPU.
For embedded devices, the difference is bigger: VPIT outperforms P2B by 135\% on TX2 and by 98\% on Xavier.
This indicated that the approach of VPIT is more suited to the robotic systems, where embedded devices are the most common, and it is harder to meet real-time requirements.

\begin{table*}[]
        \caption{Memory consumption and model sizes of some of the fastest 3D SOT methods.}
        \label{tab:memory}
        \centering
        \begin{tabular}{l|cc}
            \toprule
            \textbf{Method} & \textbf{Peak Allocated CUDA Memory} & \textbf{Parameters}  \\
            PTT-Net \cite{shan2023ptt} & \textbf{56.2 MiB} & 4903k  \\
            P2B \cite{qi2020p2b} & 433 MiB & 1339k \\
            VPIT (Ours) & 170.9 MiB & \textbf{149k} \\
            \bottomrule
        \end{tabular}
\end{table*}

Table \ref{tab:memory} compares the memory consumption and parameter count of VPIT with those of PTT-Net and P2B. VPIT with 1 block uses at least 10 times fewer parameters, but PTT has the lowest GPU memory usage. For computing GPU memory consumption, we use a tool from PyTorch that allows to see peak allocated CUDA memory, excluding memory needed for CUDA context and caching memory allocation.

\subsection{Real-time evaluation}\label{sec:experiments-real-time}
Application of 3D tracking methods is usually done for robotic systems that do not use high-end GPUs due to their high power consumption, but instead apply the computations on embedded devices, such as TX2 or Xavier. 
Following \cite{li2020towards}, we implement a predictive real-time benchmark.
Given a time step $\tau$, the set of inputs $S_{\mathrm{in}}$ that is visible to the model is limited by those inputs that appeared before $\tau$, i.e., $S_{\mathrm{in}} = (x_i, \tau_i | i \leq \tau)$, where $(x_i, \tau_i)$ is a pair of an input frame and a corresponding time step.
The processing time of the model is higher than zero, which means that the resulting prediction $p_i$ at time step $\hat{\tau}_i$ cannot be compared to the label $y_i$ from the same frame, and therefore, for each label $y_i$ from the dataset, it will be compared to the latest prediction $p_{l_{\mathrm{pr}}(i)}$.
The predictive real-time error $E_{\mathrm{pr}}$, based on a regular error $E$, is computed for each ground truth label $y_i$ as follows:
\begin{align}
\begin{split}
    &E_{\mathrm{pr}}(y_i) = E(y_i, p_{l_{\mathrm{pr}}(i)}), \\
    &l_{\mathrm{pr}}(i) = \underset{j}{\mathrm{argmax}} \:\: \hat{\tau}_j \leq \tau_i.
\end{split}
\end{align}
\begin{center}
    \begin{table*}[]
        \caption{Evaluation of some of the fastest methods for 3D Car tracking on KITTI dataset in real-time settings without the predictive requirement. The evaluation is performed with official implementations on embedded GPU platforms for 10 and 20 Hz Lidars (Data FPS). Frame drop represents the percentage of frames that could not be processed due to the model's latency. Regular represents evaluation without real-time requirements.}
        \label{tab:results-realtime-nopred}
        \centering
        \resizebox{1\columnwidth}{!}{%
        \begin{tabular}{l|ccccccccccc}
            \toprule
            \textbf{Method} & \textbf{Data FPS} & \multicolumn{3}{c}{\textbf{Success (non-predictive)}} & \multicolumn{3}{c}{\textbf{Precision (non-predictive)}} & \multicolumn{2}{c}{\textbf{FPS}} & \multicolumn{2}{c}{\textbf{Frame drop}} \\
            & & Regular & TX2 & Xavier & Regular & TX2 & Xavier & TX2 & Xavier & TX2 & Xavier\\
            \midrule
            P2B \cite{qi2020p2b} & 10 & 56.20 & 21.90 & 36.50 & 72.80 & 21.70 & 42.40 & 6.17 & 10.07 & 37.41\% & 7.00\% \\
            PTT-Net \cite{shan2023ptt} & 10 & \textbf{67.80} & 29.50 & \textbf{63.60} & \textbf{81.80} & 30.00 & \textbf{75.10} & 6.90 & 12.38 & 28.21\% & 0.81\% \\
            VPIT (Ours) & 10 & 50.49 & \textbf{50.31} & 50.49 & 64.53 & \textbf{64.08} & 64.53 & \textbf{14.31} & \textbf{20.55} & \textbf{0.68\%} & \textbf{0.00\%} \\
            \midrule
            P2B \cite{qi2020p2b} & 20 & 56.20 & 10.90 & 16.70 & 72.80 & 7.90 & 15.0 & 5.54 & 9.61 & 70.57\% & 51.56\% \\
            PTT-Net \cite{shan2023ptt} & 20 & \textbf{67.80} & 17.90 & 26.50 & \textbf{81.80} & 15.70 & 26.60 & 6.61 & 11.88 & 64.14\% & 38.95\% \\
            VPIT (Ours) & 20 & 50.49 & \textbf{38.96} & \textbf{47.70} & 64.53 & \textbf{45.50} & \textbf{59.87} & \textbf{14.37} & \textbf{20.06} & \textbf{30.17\%} & \textbf{2.38\%} \\
            \bottomrule
        \end{tabular}%
        }
    \end{table*}
\end{center}
\begin{center}
    \begin{table*}[]
        \caption{Evaluation of some of the fastest methods for 3D Car tracking on KITTI dataset in real-time settings with the predictive requirement. The evaluation is performed with official implementations on embedded GPU platforms for 10 and 20 Hz Lidars (Data FPS). Frame drop represents the percentage of frames that could not be processed due to the model's latency. "Regular" represents evaluation without real-time requirements}
        \label{tab:results-realtime-pred}
        \centering
        \resizebox{1\columnwidth}{!}{%
        \begin{tabular}{l|ccccccccccc}
            \toprule
            \textbf{Method} & \textbf{Data FPS} & \multicolumn{3}{c}{\textbf{Success (predictive)}} & \multicolumn{3}{c}{\textbf{Precision (predictive)}} & \multicolumn{2}{c}{\textbf{FPS}} & \multicolumn{2}{c}{\textbf{Frame drop}} \\
            & & Regular & TX2 & Xavier & Regular & TX2 & Xavier & TX2 & Xavier & TX2 & Xavier\\
            \midrule
            P2B \cite{qi2020p2b} & 10 & 56.20 & 19.10 & 30.30 & 72.80 & 17.80 & 33.70 & 6.17 & 10.07 & 36.31\% & 6.79\% \\
            PTT-Net \cite{shan2023ptt} & 10 & \textbf{67.80} & 24.90 & \textbf{50.70} & \textbf{81.80} & 23.40 & 59.00 & 7.07 & 12.26 & 26.15\% & 0.90\% \\
            VPIT (Ours) & 10 & 50.49 & \textbf{45.82} & 46.68 & 64.53 & \textbf{57.76} & \textbf{59.28} & \textbf{14.61} & \textbf{20.55} & \textbf{0.57\%} & \textbf{0.00\%} \\
            \midrule
            P2B \cite{qi2020p2b} & 20 & 56.20 & 9.10 & 13.50 & 72.80 & 6.00 & 10.90 & 5.54 & 9.47 & 69.57\% & 50.31\% \\
            PTT-Net \cite{shan2023ptt} & 20 & \textbf{67.80} & 15.40 & 23.20 & \textbf{81.80} & 13.10 & 21.50 & 6.67 & 12.06 & 62.49\% & 37.35\% \\
            VPIT (Ours) & 20 & 50.49 & \textbf{34.00} & \textbf{41.39} & 64.53 & \textbf{36.61} & \textbf{49.36} & \textbf{14.40} & \textbf{20.77} & \textbf{29.41\%} & \textbf{1.56\%} \\
            \bottomrule
        \end{tabular}%
        }
    \end{table*}
\end{center}

As shown by \cite{li2020towards}, the existing model can be improved for a predictive real-time evaluation by predicting the position of the object at a frame $i$ and then predicting its movement during the time period between the frames $i$ and $i + 1$ using a Kalman Filter \cite{1960kalmanfilter} or any other similar method.
Such optimization can be applied to any of the methods that we compare, and therefore, to eliminate the factor of ``predictive errors" where the model's error in prediction for the current frame is more leaned towards the prediction for the next frame, we introduce a non-prediction real-time benchmark.
This benchmark effectively shifts all labels one frame forward, allowing the methods that are faster than the data FPS to be evaluated in the same way, as in a regular evaluation protocol. The only difference between the non-predictive and predictive benchmarks is in the way the latest prediction is defined:
\begin{equation}
    l_{\mathrm{npr}}(i) = \underset{j}{\mathrm{argmax}} \:\: \hat{\tau}_j \leq \tau_{i+1}.
\end{equation}

We evaluate some of the fastest methods 3D SOT methods (P2B, PTT, VPIT) on embedded devices with both non-predictive (Table \ref{tab:results-realtime-nopred}) and predictive (Table \ref{tab:results-realtime-pred}) benchmarks.
Since the tested methods except VPIT lose most of their performance in the real-time benchmark scenario, and other methods are $1.5$ times slower or more, the evaluation of those methods will result in the complete inability to track objects and can be omitted.
We select Data FPS to be either 10 or 20, representing the most popular 10 and 20 Hz Lidars.
The 10Hz Lidar is the easier case, as for the method to be real-time, it needs to sustain FPS, higher than 10, compared to 20 for the 20 Hz Lidar.
During the evaluation, we compute how many frames could not be processed due to the model's latency and represent it as a Frame drop percentage.
\begin{figure}[!t]
\begin{center}
    \includegraphics[width=0.9\linewidth]{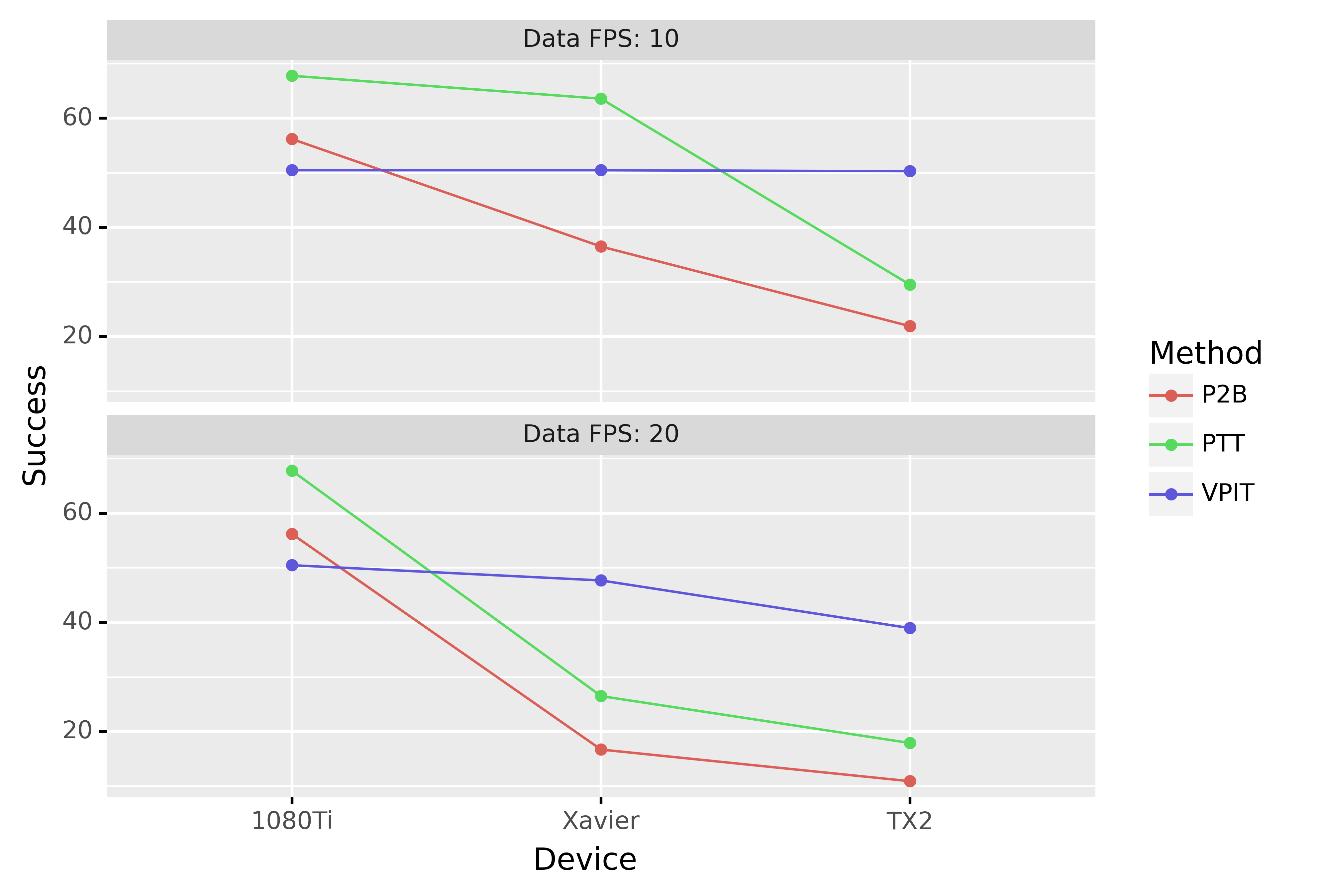}
    \caption{Evaluation of some of the fastest methods models with real-time requirements on embedded devices and a desktop GPU. Data FPS represents 10 and 20 Hz Lidars. Devices are sorted in descending order by their computational power.}
    \label{fig:realtime}
\end{center}
\end{figure}

Only VPIT on Xavier (more powerful embedded device) for a 10Hz Lidar could not suffer from any frame drop, resulting in the regular Success and Precision values for a non-predictive benchmark and suffering $3.81$ Success and $5.25$ Precision for a predictive benchmark.
The highest Frame drop of $60-70\%$ is seen for P2B and PTT on TX2 for a 20Hz Lidar.
In this case, Success values of P2B and PTT are $6$ and $4$ times lower than during the regular evaluation, respectively, which indicates that these methods cannot work under the aforementioned conditions.
For every evaluation case, except the Xavier with a 10Hz Lidar, VPIT outperforms P2B and PTT in Success, Precision, FPS and Frame drop, but PTT still maintains the best Success and Precision on Xavier with a 10Hz Lidar.
As can be seen from Fig. \ref{fig:realtime}, the Success drop of VPIT is the smallest, while P2B and PTT lose most of their tracking accuracy on TX2 for any Lidar and on Xavier for a 20Hz Lidar.

Evaluation of the other methods, presented in Table \ref{tab:results-reported}, is not needed, as their FPS is $1.5$ or more times slower than the FPS of P2B and PTT, which will result in a complete loss of Success and Precision on any of the real-time benchmarks that were discussed.

\begin{figure*}[]
\begin{center}
    \includegraphics[width=1\linewidth]{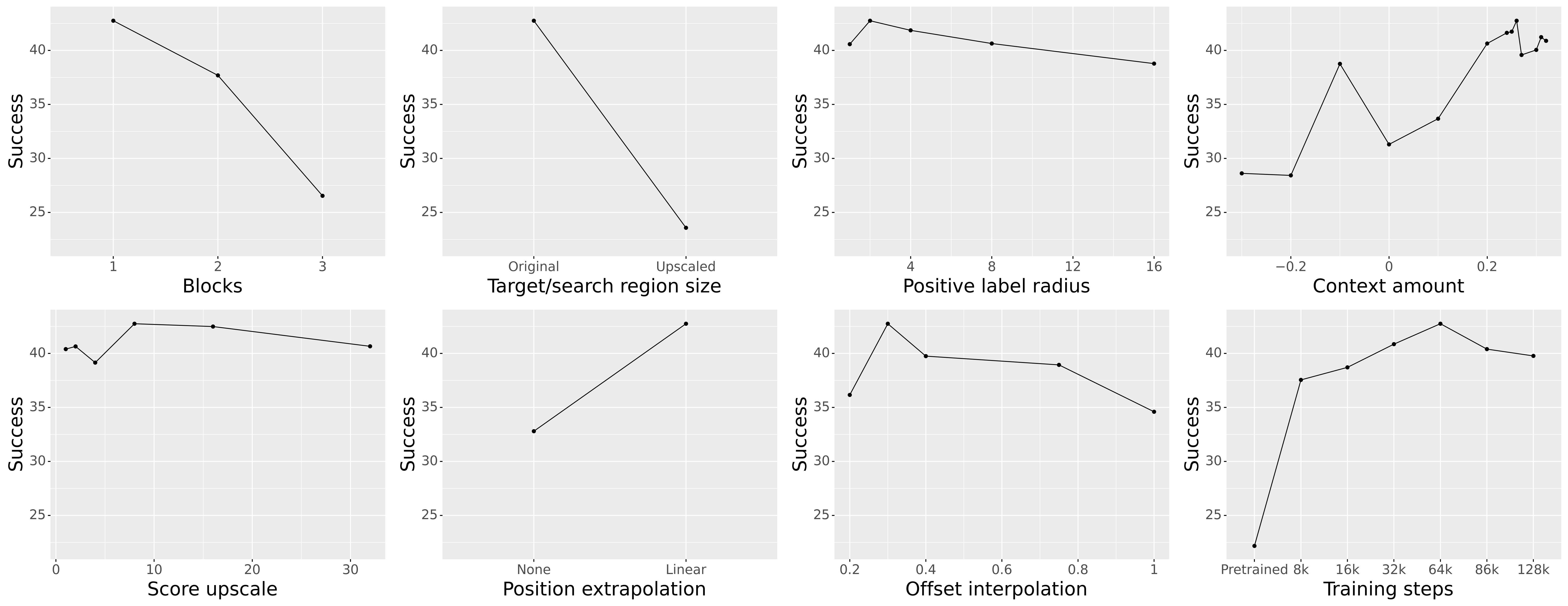}
    \caption{Influence of different hyperparameter values on Success metric on validation KITTI Tracking subset. }
    \label{fig:ablation}
\end{center}
\end{figure*}

\subsection{Ablation study}\label{sec:experiments-ablation}

We performed experiments to determine the influence of different hyperparameters on the Success and Precision metrics.
We used KITTI Tracking tracks $10-11$ for validation and $0-9$, $12-18$ for training.
The influence of selected hyperparameters on Success and Precision metrics is similar, and therefore we present only the effect on the Success metric on the validation subset in Fig. \ref{fig:ablation}.

Decreasing the number of feature blocks from the Feature Generation Network, leads to better Success values. This can be due to an increasing receptive field with each new block that leads to over smoothing of the resulting feature maps and makes it harder to define a precise position of an object of interest.
In addition to the best model accuracy, using one block leads to the smallest model size. The VPIT model with one block contains $149k$ parameters, while with 2 blocks the number of parameters is increased to $961k$, and with all 3 blocks the size of the model is $4208k$ parameters.

Next, we follow SiamFC \cite{bertinetto2016siamfc} and use target and search upscaling with $\iota_t = (127, 127)$ and $\iota_s = (255, 255)$.
This results in constant-sized target and search pseudo-images, corresponding feature and score maps, which allows implementing mini-batch training and reduce the training time.
However, such an approach does not work well for voxel pseudo images, achieving $81\%$ less Success than when using original sizes ($\iota_t = (0, 0)$ and $\iota_s = (0, 0)$).

Furthermore, following SiamFC \cite{bertinetto2016siamfc}, positive values of the context amount parameter $c$ result in square target and search regions.
In contrast to it, we also use negative $c$ values to add independent context to each dimension of the region, keeping its aspect ratio.
Positive context leads to better Success with the maximum at $c = 0.26$, which is $10\%$ higher than the result for $c = -0.1$.

Regarding the positive label radius, $r=2$ results in the best training procedure, but this hyperparameter has a low influence on the final Success value, as well as the target feature merge scale $m_{\mathrm{tf}}$, window influence $\eta$, score upscale $u_M$, search region scale $\sigma_s$, rotation step $\psi_\alpha$ and rotation penalty $\Lambda$.

Next, linear position extrapolation leads to a $30\%$ increase in Success compared to a conventional search region placement and no directional penalty.
With offset interpolation $\omega = 0.3$, the Success value increases by $24\%$, compared to the direct application of the model's prediction ($\omega = 1$).

Finally, a pretrained PointPillars backbone creates features that are capable of differentiating between target class and other object types, but those features cannot be used effectively to differentiate between different objects from the same class.
This serves as a good starting point for the SOT task, and then training for $64000$ steps results in the best model for selected hyperparameters.

\subsection{Limitations and future work}
VPIT operates on Birds-Eye-View (BEV) data and shares all the limitations of BEV methods, which includes the computation of $z$ values of the final boxes.
Due to the nature of current 3D SOT datasets, tracking is usually done in $x$ and $y$ coordinates with small or almost no changes in $z$ values.
For this reason, VPIT reuses $z$ position from the initial object position. A straightforward extension would be to regress it by a special subnetwork to perform fine $z$-tracking too.
One of the limitations of the method lies in how multi-rotation search works.
The precision of rotations is defined by the number of search angles that are checked and the difference between them. Therefore, to increase the precision, one needs to use more search regions, which has a negative impact on the memory and inference speed of the model.
Similar to any other 3D SOT method, when the density of point cloud is too small due to high object distance, the method cannot provide as precise position as for close objects, and can lose the object completely given the instability of points at far distances.

\section{Conclusions}\label{sec:conclusions}
In this work, we proposed a novel method for 3D SOT called VPIT.
We focus on stepping out of the point-based approaches for this problem and studying methods to use structured data for 3D SOT.
We formulate a lightweight method that uses PointPillars' pseudo images as a search space and applies a Siamese 2D-like approach on these pseudo images to find both the position and rotation of the object of interest.
Experiments show that VPIT is the fastest method, and it achieves competitive Precision and Success values.
Moreover, we implement a real-time evaluation protocol for 3D single object tracking and use it for evaluation of the fastest methods on embedded devices, that are the most popular choice for robotic systems. Results showcase that other methods are less suited for such devices and lose most of their ability to track objects due to a high latency of predictions.
The proposed method allows for an easy adaptation of 2D tracking ideas to a 3D SOT task while keeping the model lightweight.

\section*{Acknowledgements}
This work has received funding from the European Union’s Horizon 2020 research and innovation programme under grant agreement No 871449 (OpenDR). This publication reflects the authors’ views only. The European Commission is not responsible for any use that may be made of the information it contains.

\section*{Data Availability Statement}
The KITTI \cite{2012kitti} dataset used in this work is publicly available.

\section*{Conflict of interest}
The authors declare that they have no conflicts of interest.

\bibliography{bibliography}

\end{document}